\documentclass[article, 10pt]{acmart}

\setcopyright{none}

\renewcommand\footnotetextcopyrightpermission[1]{}

\AtBeginDocument{%
  }

\setcopyright{none}
\renewcommand\footnotetextcopyrightpermission[1]{}
\usepackage{enumitem}
\usepackage{xurl}  
\usepackage{balance}  
\usepackage{tikz}
\usepackage{adjustbox}
\newcommand{\mx}{\textsc{Mod-X}}

\usetikzlibrary{positioning,fit,shapes.misc}

\usepackage{listings}
\begin{document}


\title{{\mx}: A Modular Open Decentralized eXchange Framework\\proposal for Heterogeneous Interoperable Artificial Intelligence Agents}

\author{Georgios Ioannides}
\email{gioannid@alumni.cmu.edu}
\affiliation{%
  \institution{James Silberrad Brown Center for Artificial Intelligence}
  \institution{Carnegie Mellon University}
  \institution{Amazon GenAI$^{*}$}
  \country{USA}
}

\author{Christos Constantinou}
\email{christos.constantinou@bristol.ac.uk}
\affiliation{%
  \institution{University of Bristol}
  \institution{James Silberrad Brown Center for Artificial Intelligence}
  \institution{Amazon GenAI$^{*}$}
  \country{USA}
}

\author{Vinija Jain}
\email{hi@vinija.ai}
\affiliation{%
  \institution{James Silberrad Brown Center for Artificial Intelligence}
  \country{USA}
}

\author{Aman Chadha}
\email{hi@aman.ai}
\affiliation{%
  \institution{James Silberrad Brown Center for Artificial Intelligence}
  \institution{Amazon GenAI$^{*}$}
  \country{USA}
}

\author{Aaron Elkins}
\email{aelkins@sdsu.edu}
\affiliation{%
  \institution{James Silberrad Brown Center for Artificial Intelligence}
  \country{USA}
}


\begin{abstract}
  As Artificial Intelligence systems evolve from monolithic models to ecosystems of specialized agents, the need for standardized communication protocols becomes increasingly critical. This paper introduces {\mx} (\textbf{M}odular \textbf{O}pen \textbf{D}ecentralized e\textbf{X}change), a novel architectural framework proposal for agent interoperability that addresses key limitations of existing protocols. Unlike current approaches, {\mx} proposes a layered architecture with a Universal Message Bus, thorough state management, translation capabilities, and blockchain-based security mechanisms. We present {\mx}'s architecture, compare it with existing protocols, and demonstrate its application through a worked example how it enables integration between heterogeneous specialist agents (agents with different architectures, vendors, capabilities, and knowledge representations—including rule-based systems, neural networks, symbolic reasoning engines, and legacy software with agent wrappers). {\mx}'s key innovations include a publish-subscribe communication model, semantic capability discovery, and dynamic workflow orchestration—providing a framework that bridges theoretical formalism with practical implementation. This architecture addresses the growing need for truly decentralized, interoperable agent ecosystems that can scale effectively without the need for central coordination.
\end{abstract}

\ccsdesc[500]{Information systems~Information retrieval}
\ccsdesc[300]{Computing methodologies~Multi-agent systems}
\ccsdesc[300]{Computing methodologies~Distributed artificial intelligence}

\keywords{AI agents, agent communication protocols, interoperability, decentralized exchange}


\maketitle

\pagestyle{plain}
\thispagestyle{empty} 

\renewcommand{\thefootnote}{\fnsymbol{footnote}}
\footnotetext[1]{Work does not relate to position at Amazon.}
\section{Introduction}

Artificial Intelligence (AI) is undergoing a paradigm shift from monolithic, general-purpose models toward ecosystems of specialized, interoperable agents. This transition presents significant challenges in enabling effective communication and collaboration among diverse AI agents built by different vendors, each with varying architectures and capabilities. Current approaches to agent communication often fall into one of two categories: theoretical frameworks with limited practical implementation (like FIPA-ACL \cite{fipa2002acl}) or pragmatic but specialized solutions for specific ecosystems (like LangChain's Agent Protocol \cite{langchain2024agent}).

The emerging initiatives such as Agent-to-Agent (A2A) protocol \cite{google2024a2a} and Model Context Protocol \cite{anthropic2024mcp} (MCP) represent important steps toward standardization, but they focus on different aspects of the agent communication problem. A2A enables agent-to-agent interactions without exposing internal state, while MCP standardizes how AI applications access external tools and data sources. However, neither protocol provides an elaborate solution for truly decentralized, modular agent ecosystems that can scale effectively.

Agents' primary use cases are research and summarization tasks, which fall under information retrieval \cite{langchain2024stateofai}. Current agent communication protocols significantly limit the autonomy and effectiveness of information retrieval tasks by lacking sophisticated mechanisms for agent-to-agent coordination. The proposed {\mx} approach can improve the overall autonomy of information retrieval through better agent-to-agent communication mechanisms.

In this paper, we introduce {\mx} (Modular Open Decentralized eXchange), a proposed architectural framework designed to enable integration of heterogeneous specialist AI agents. Our contributions include:

\begin{enumerate}
    \item A thorough analysis of existing agent communication protocols/frameworks, highlighting their strengths and limitations
    \item The introduction of {\mx}'s layered architecture with a Universal Message Bus, translation layer, state management, and security mechanisms
    \item A detailed explanation of {\mx}'s key components, including capability-based discovery, dynamic workflow orchestration, and blockchain-based verification
    \item A worked example demonstrating how {\mx} facilitates complex multi-agent interactions in practical scenarios
\end{enumerate}

Unlike existing approaches that focus on specific aspects of agent communication—whether enterprise security (A2A), tool integration (MCP), or distributed coordination (NANDA \cite{nanda2024mit})—{\mx} provides a translation and interoperability layer that can bridge these diverse protocol families while maintaining their individual strengths.

\section{Related Work}

Agent communication protocols have evolved from theoretical frameworks to practical implementations, yet existing approaches provide limited support for effective information retrieval tasks from multiple heterogeneous sources that require sophisticated coordination between these heterogeneous agents. Based on recent surveys \cite{yang2024survey}, we analyze existing approaches in different architectural categories as detailed in Table~\ref{tab:protocol_survey}.

\subsection{Evolution and Technical Challenges}

Early theoretical frameworks like FIPA-ACL established formal foundations using speech act theory—the idea that when agents send messages, they are performing actions like making requests or sharing beliefs, not just exchanging data. For example, when Agent A sends "I request that you book a flight," this message actually performs the action of making a request. However, these theoretical frameworks face a fundamental verification problem: there's no way to verify whether an agent's expressed beliefs and intentions align with their actual internal mental states. You can't "peek inside an agent's programming" to confirm if it truly believes what it claims.

The current protocol landscape shows four distinct architectural approaches addressing different communication challenges. Context-oriented protocols solve the "agent needs tools" problem by standardizing how agents access external resources. The Model Context Protocol (MCP) provides a universal standard for connecting LLM agents to external data, tools, and services using a client-server architecture where the agent (Host) connects to multiple Clients, each managing specific Resources through dedicated Servers. This decouples tool invocation from LLM responses, reducing data leakage risks while enabling standardized context acquisition. Similarly, agent.json \cite{wildcard2025agent} creates machine-readable contracts for websites to declare AI-compatible interfaces, built on OpenAPI standards to enable websites to publish structured JSON files at `/.well-known/agents.json` that define workflows, authentication schemes, and data dependencies, making web services more accessible to AI agents.

General-purpose inter-agent protocols solve the "agents need to talk to each other" problem through various approaches. The Agent Network Protocol \cite{aiengr2025agent} (ANP) aims to create an "Internet of Agents" using three layers: Identity (W3C DID for decentralized authentication), Meta-Protocol (natural language protocol negotiation), and Application Protocol (standardized agent discovery and interaction). The Agent2Agent Protocol (A2A) enables enterprise-grade agent collaboration through HTTP-based messaging with JSON-RPC 2.0 format, emphasizing async-first architecture supporting long-running tasks, multi-modal communication, and enterprise security requirements while maintaining "opaque execution" where agents don't share internal thoughts or tools. The Agent Interaction \& Transaction Protocol \cite{aitp2025near} (AITP) focuses on secure cross-trust-boundary communication using blockchain technology for identity verification and value exchange between autonomous agents. Both the Agent Communication Protocol \cite{acomp2025ibm} (AComP) and Agent Connect Protocol \cite{langchain2025aconp} (AConP) provide standardized interfaces for agent invocation and configuration, focusing on practical deployment needs. Agora \cite{oxford2024agora} addresses the "Agent Communication Trilemma"—balancing versatility, efficiency, and portability by enabling agents to autonomously negotiate communication protocols using Protocol Documents (PDs), allowing dynamic adaptation between structured protocols, LLM routines, and natural language based on communication frequency and context.

Domain-specific inter-agent protocols solve specialized interaction challenges across three main areas. For human-agent interaction, the PXP Protocol \cite{bits2024pxp} enables bidirectional intelligible communication using finite-state machines with RATIFY/REFUTE/REVISE/REJECT message tags, while LOKA \cite{cmu2025loka} provides decentralized identity and ethical consensus mechanisms for AI governance. For robot-agent interaction, CrowdES \cite{gist2025crowdes} generates realistic crowd behavior for robot navigation environments, and Spatial Population Protocols solve distributed localization problems among anonymous robots through geometric consensus algorithms. For system-agent interaction, the Language Model Operating System \cite{lmos2025eclipse} (LMOS) creates an "Internet of Agents" infrastructure with three layers for agent discovery, communication protocol negotiation, and identity management, while Agent Protocol defines framework-agnostic standards for agent lifecycle management (starting, stopping, monitoring) using OpenAPI specifications.

A key insight from recent research is that context-oriented and inter-agent protocols may gradually converge. Tools can be viewed as "low-autonomy agents," while agents can function as "high-autonomy tools." This suggests future protocols may need to handle both paradigms uniformly—where an agent accessing a database tool through MCP could be equivalent to an agent collaborating with a specialized data-analysis agent through A2A. The rapid evolution is evident: MCP launched in November 2024 initially lacked HTTP support, but by early 2025 had added HTTP Server-Sent Events, authentication, and HTTP Streaming capabilities, mirroring the TCP/IP to HTTP evolution in internet protocols.

The analysis reveals fundamental gaps preventing effective information retrieval coordination:
\begin{enumerate}
    \item \textbf{Semantic Fragmentation}: protocols use incompatible vocabularies and ontologies, preventing agents from discovering equivalent capabilities across different implementations.
    \item \textbf{State Management Conflicts}: stateless approaches (A2A) cannot maintain research context across multi-step workflows, while stateful systems (AutoGen \cite{microsoft2025autogen}, AComP) lack distributed coordination mechanisms.
    \item \textbf{Security-Interoperability Tension}: enterprise security models (OAuth/JWT) create authentication silos that prevent cross-organizational agent collaboration essential for extensive research tasks.
    \item \textbf{Limited Cross-Domain Coordination}: domain-specific optimizations (IoT, robotics, blockchain) cannot generalize to information retrieval scenarios requiring diverse agent types.
\end{enumerate}

\subsection{Fundamental Architectural Gap}

No existing approach provides the combination of universal semantic translation, distributed state coordination, and cross-protocol interoperability required for next-generation agent ecosystems focused on information retrieval. Current protocols optimize for specific communication patterns or deployment contexts but lack the architectural flexibility to bridge heterogeneous agent implementations while maintaining the sophisticated coordination mechanisms necessary for complex research and summarization workflows.
\section{{\mx}: Architecture and Components}

\begin{figure}[t]
\centering
\includegraphics[width=0.35\linewidth, trim=3.9cm 0cm 3.9cm 0.5cm]{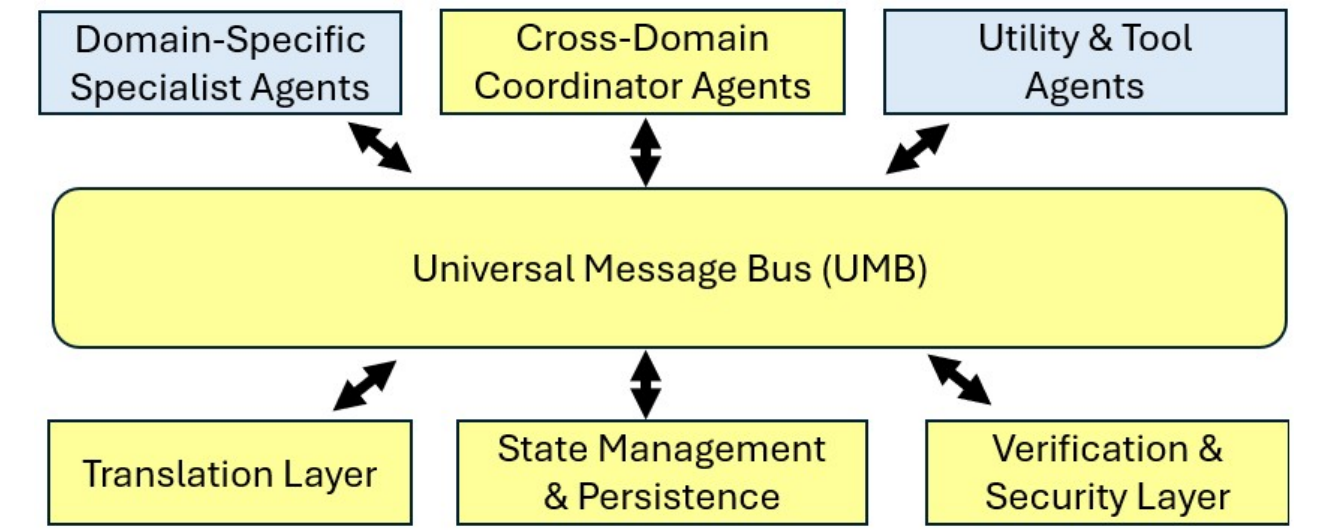}
\caption{{\mx} layered architecture showing different agent types, Universal Message Bus, and supporting layers}
\label{fig:modx-architecture}
\vspace{-1em}
\end{figure}

The {\mx} (Modular Open Decentralized eXchange) framework proposal is illustrated in Figure \ref{fig:modx-architecture} which builds on lessons from existing protocols to provide a proposed implementation architecture for truly interoperable agent systems. The framework is designed to enable integration between heterogeneous specialist agents through a layered architecture that addresses key challenges for enhanced information retrieval (between agents) in agent communication.

\subsection{Architecture Overview}

{\mx} proposes a layered architecture with clear separation of concerns.

This architecture supports diverse agent types that can dynamically join and leave the ecosystem, with the Universal Message Bus providing the core communication infrastructure.

\subsection{Core Architecture}

\subsubsection{Universal Message Bus (UMB)}
The communication backbone that connects all three agent types in the ecosystem:

\begin{itemize}
    \item \textbf{Domain-Specific Specialist Agents}: Handle specific domain tasks (e.g., flight booking, weather forecasting)
    \item \textbf{Cross-Domain Coordinator Agents}: Orchestrate multi-domain workflows  
    \item \textbf{Utility \& Tool Agents}: Provide supporting services (authentication, logging, translation)
\end{itemize}

The UMB provides:
\begin{itemize}
    \item \textbf{Decentralized routing}: Agents communicate through topic-based publish-subscribe \cite{eugster2003many, tarkoma2012publish}, eliminating the need for direct point-to-point connections. For example, when a weather monitoring agent detects flight delays due to storms, it publishes this information to the 'flight-disruption' topic, automatically notifying all subscribed travel planning agents without needing to know their specific identities or locations. Similarly, a budget tracking agent can subscribe to 'expense-transaction' topics to monitor spending across multiple booking agents simultaneously, enabling real-time budget enforcement across the entire ecosystem.
    \item \textbf{Reliable delivery}: WebSocket connections with HTTP fallback ensure message delivery across network conditions
\end{itemize}

Unlike MCP's client-server model or A2A's direct agent connections, UMB enables many-to-many interactions where any agent can communicate with any other agent without prior configuration, regardless of whether they are domain specialists, coordinators, or utility agents.

\subsection{Supporting Infrastructure Layers}

\subsubsection{Translation Layer}
Operates beneath the UMB to solve the interoperability challenge of connecting agents built with different technologies and vocabularies:

\begin{itemize}
    \item \textbf{Semantic capability matching}: Agents describe their functions using vector embeddings and ontological links, allowing discovery based on meaning rather than exact keyword matches
    \item \textbf{Automatic message translation}: Converts between different agent message formats and data representations in real-time. For instance, when a European hotel booking agent describes room types as 'Standard/Superior/Deluxe' but an American travel agent expects 'Economy/Business/Premium', the translation layer automatically maps these equivalent concepts using semantic embeddings. Currency conversions, date format differences (DD/MM/YYYY vs MM/DD/YYYY), and measurement units (kilometers vs miles) are handled transparently without requiring agents to be aware of these differences.
\end{itemize}

This addresses the fundamental problem where agents developed independently use different vocabularies for the same capabilities. The UMB addresses the classic API integration challenge of connecting multiple applications with multiple systems without requiring custom point-to-point integrations \cite{pautasso2008restful}, enabling standardized communication patterns that scale efficiently.

\subsubsection{State Management \& Persistence}
Manages how agents operate and share information in the distributed system:

\begin{itemize}
    \item \textbf{Containerized deployment}: Agents run in standardized containers regardless of their internal technology (e.g. Python, JavaScript, neural networks, rule engines)
    \item \textbf{Contextual state sharing}: Maintains shared context for complex multi-step interactions while preserving agent autonomy
    \item \textbf{Distributed consistency}: Resolves conflicts when multiple agents modify shared information. For example, during a complex business trip booking, the flight agent temporarily shares seat preferences and frequent flyer status with the hotel agent to maintain service level consistency, but this shared context expires once the booking is complete. Multiple agents coordinating an emergency rebooking can access shared context about the traveler's constraints and preferences without permanent access to personal data.
\end{itemize}

This enables complex collaborations that may require maintaining context across multiple agent interactions, which existing stateless protocols cannot support. State management adapts distributed consensus algorithms \cite{ongaro2014raft, lamport1998paxos} and container orchestration patterns \cite{zhou2021container_real} proven in production cloud environments.

\subsection{Cross-Domain Coordination}

\subsubsection{Workflow Orchestration}
Primarily handled by Cross-Domain Coordinator Agents to coordinate complex multi-agent processes that go beyond simple message exchanges:

\begin{itemize}
    \item \textbf{Declarative workflows}: Define multi-step processes as graphs showing which agents need to run and in what order
    \item \textbf{Parallel execution}: Automatically identifies and runs independent operations simultaneously
    \item \textbf{Failure recovery}: When an agent fails, the system can retry, substitute alternative agents, or roll back completed work
\end{itemize}

This enables sophisticated patterns like gathering information from multiple sources, building consensus among agents, and multi-stage decision refinement.

\subsection{Security \& Trust}
\subsubsection{Verification \& Security Layer}
Ensures secure and trustworthy interactions across all agent types in the decentralized ecosystem: (1) \textbf{Cryptographic identity}: Each agent has a verifiable digital identity that cannot be forged and all agent communications are recorded in tamper-proof logs, (2) \textbf{Reputation tracking}: Agents build trust scores based on successful interactions and honest behavior. This blockchain-based approach provides stronger security guarantees than traditional OAuth/JWT systems, addressing the unique trust challenges of decentralized agent networks through reputation tracking mechanisms \cite{calvaresi2018mas}. Blockchain operations introduce computational overhead \cite{zhang2021blockchain_ai, calvaresi2018mas}. To address this, {\mx} implements a tiered security model: (1) \textbf{High-value operations} use blockchain such as for agent registration, financial transactions, reputation updates, and security policies; (2) \textbf{Routine communications} employ lightweight cryptographic verification such as for messages, queries, and status updates. This selective on-chain/off-chain approach significantly reduces overhead while maintaining security guarantees for critical operations \cite{calvaresi2018mas}, using digital signatures with periodic batch anchoring to blockchain.

\subsection{Implementation Status and Feasibility}

MOD-X represents a conceptual architecture that builds upon established technologies and proven design patterns. While the complete integrated system has not been implemented, each component leverages well-established foundations:

\begin{itemize}
    \item \textbf{Universal Message Bus}: Built on proven publish-subscribe architectures widely deployed in production systems \cite{eugster2003many}
    \item \textbf{Translation Layer}: Leverages recent advances in embedding space alignment \cite{jha2025harnessing}
    \item \textbf{State Management}: Adapts distributed consensus algorithms \cite{ongaro2014raft, lamport1998paxos} and container orchestration patterns \cite{netto2017state, zhou2021container} proven in production cloud environments
    \item \textbf{Security Layer}: Applies blockchain-based trust mechanisms increasingly validated in multi-agent systems \cite{calvaresi2018mas} and distributed computing environments
\end{itemize}

\section{Comparison with Existing Protocols}

{\mx} differs from existing protocols in several important dimensions, addressing key limitations while building on their strengths.

\subsection{Agent Diversity in {\mx}}
A key strength of {\mx} is its ability to integrate diverse agent types regardless of their underlying implementation. While recent attention has focused on Large Language Model (LLM) based agents, {\mx}'s architecture is fundamentally technology-agnostic. Agents in the {\mx} ecosystem can be implemented using any technology, including \textbf{traditional AI systems} (rules, logic programming, symbolic reasoning), \textbf{machine learning models} (neural networks, decision trees, statistical systems), \textbf{expert systems} (domain-specific knowledge representation), \textbf{legacy software systems} (with agent wrappers), \textbf{IoT devices and physical sensors} (environmental data), and \textbf{human-in-the-loop systems} (human-augmented AI capabilities).
This technology-agnostic approach enables {\mx} to serve as a universal connector across the entire spectrum of AI technologies, from classical symbolic systems to modern neural approaches, and even non-AI systems that can benefit from agent-based interfaces. The Translation Layer plays a crucial role in this regard, enabling communication between heterogeneous implementations by transforming message formats and aligning semantic representations. As a result, {\mx} can bridge diverse AI generations and paradigms within a single coherent ecosystem.

\begin{table*}[t]
\centering
\tiny
\adjustbox{width=\textwidth}{
\begin{tabular}{lccccccc}
\toprule
Protocol & Year & Communication Pattern & Security Model & State Management & Key Innovation & Target Domain & Maturity \\
\midrule
\multicolumn{8}{l}{\textbf{Agent-to-Resource Integration Protocols}} \\
MCP \cite{anthropic2024mcp} & 2024 & RPC Client-Server & OAuth & Server-side & Simplified tool integration & Tool integration & Deployed \\
agent.json \cite{wildcard2025agent} & 2025 & HTTP Well-known & Web standards & Stateless & Automatic web integration & Web information & Draft \\
\midrule
\multicolumn{8}{l}{\textbf{General-Purpose Agent-to-Agent Communication}} \\
A2A \cite{google2024a2a} & 2025 & JSON-RPC 2.0 & OAuth/JWT & Stateless & Enterprise security model & Enterprise agents & Deployed \\
ANP \cite{anp2024community} & 2024 & P2P JSON-LD & DID/Verifiable Creds & Distributed & Decentralized identity (DID) & Decentralized agents & Prototype \\
AITP \cite{aitp2025near} & 2025 & HTTP/Blockchain & Crypto signatures & Blockchain state & Economic incentives & Economic agents & Concept \\
AComP \cite{acomp2025ibm} & 2025 & OpenAPI & Enterprise auth & Stateful & Enterprise integration & Corporate MAS & Draft \\
AConP \cite{langchain2025aconp} & 2025 & OpenAPI/JSON & API keys & Framework state & LLM orchestration & LLM orchestration & Draft \\
\midrule
\multicolumn{8}{l}{\textbf{Meta-Protocol \& System Coordination}} \\
Agora \cite{oxford2024agora} & 2024 & Protocol bridging & Protocol-specific & Meta-state & Universal protocol bridging & Protocol translation & Research \\
LMOS \cite{lmos2025eclipse} & 2025 & WoT/HTTP & DID & Distributed & Physical-digital integration & IoT-agent hybrid & Prototype \\
Agent Protocol \cite{aiengr2025agent} & 2025 & RESTful API & API authentication & Stateful sessions & Simple controller pattern & Agent controllers & Deployed \\
\midrule
\multicolumn{8}{l}{\textbf{Domain-Specialized Application Protocols}} \\
LOKA \cite{cmu2025loka} & 2025 & DECP P2P & Decentralized auth & Consensus-based & Autonomous system reliability & Autonomous systems & Research \\
PXP \cite{bits2024pxp} & 2024 & Custom protocols & User authentication & Session-based & Human-AI interaction patterns & Human-AI interaction & Research \\
CrowdES \cite{gist2025crowdes} & 2025 & Robot messaging & Physical security & Real-time state & Crowd robotics coordination & Crowd robotics & Research \\
SPPs \cite{liverpool2024spps} & 2024 & Semantic primitives & Robot protocols & Behavior state & Semantic robot coordination & Robot coordination & Research \\
\midrule
\multicolumn{8}{l}{\textbf{Foundational Theoretical Frameworks}} \\
FIPA-ACL \cite{fipa2002acl} & 2002 & Speech acts & Basic auth & Mental states & Formal communication semantics & Formal MAS & Legacy \\
\midrule
\multicolumn{8}{l}{\textbf{Development \& Orchestration Platforms}} \\
LangChain Agent \cite{langchain2024agent} & 2024 & RESTful threads & Framework auth & Persistent threads & Development simplicity & LLM applications & Deployed \\
AutoGen \cite{microsoft2025autogen} & 2025 & Event messaging & Framework security & Multi-layer state & Conversational patterns & Enterprise MAS & Deployed \\
\midrule
\multicolumn{8}{l}{\textbf{Universal Interoperability Solutions}} \\
{\mx} & 2025 & Pub/Sub + RPC & Blockchain + reputation & Contextual sharing & Cross-protocol translation & Heterogeneous agents & Research \\
\bottomrule
\end{tabular}}
\caption{Taxonomy of agent communication protocols by architectural approach and technical characteristics}
\label{tab:protocol_survey}
\end{table*}

Table~\ref{tab:protocol_survey} presents an elaborate comparison of {\mx} with major agent communication frameworks. {\mx} implements a fully decentralized publish-subscribe model with the Universal Message Bus, dynamic semantic discovery, distributed state management, extensive translation capabilities, and blockchain-based security. This unique combination enables {\mx} to support truly flexible many-to-many agent interactions without requiring central coordination or predefined connections.

\subsection{Capability Discovery}

{\mx}'s capability discovery mechanism represents a theoretical advancement from the principles of distributed cognition toward emergent semantic intelligence. This section examines the practical implementation of agent capability discovery.

\subsubsection{The Ontological Foundations of Agent Registration}

When an agent participates in the {\mx} ecosystem, it contributes to a collective intelligence through ontological self-representation. As shown in Listing~\ref{lst:aidl}, this declaration transcends mere technical specification to become a semantic identity statement.

\begin{lstlisting}[caption={Agent capability registration in AIDL format}, label={lst:aidl}]
{
  "agentId": "flight-agent-001",
  "capabilities": [{
    "name": "flightBooking",
    "version": "1.2.0",
    "semantics": {
      "ontology": "http://schema.org/Flight",
      "embedding": [0.2, 0.8, 0.1, 0.7, ...],
      "operations": ["search", "price", "book", "cancel"]
    },
    "interface": {
      "inputs": {
        "search": {
          "origin": "string",
          "destination": "string",
          "departureDate": "date",
          "returnDate": "date?",
          "passengers": "integer",
          "class": "string?"
        },
        "book": {
          "flightId": "string",
          "passengers": "array<object>",
          "paymentMethod": "object"
        }
      },
      "outputs": {
        "search": "array<object>",
        "book": "object"
      }
    }
  }]
}
\end{lstlisting}

The registration process embodies four essential ontological commitments, illustrated through our \texttt{flight-agent-001} example:

\begin{enumerate}
    \item \textbf{Epistemic Self-Awareness}: The agent explicitly declares what it knows and can do. For instance, our flight agent knows how to search, price, book, and cancel flights—but makes no claims about processing payments or managing frequent flyer programs. This self-awareness of boundaries is critical for reliable agent interactions.
    
    \item \textbf{Semantic Embedding}: The capability is represented in a multidimensional semantic space through the embedding vector \texttt{[0.2, 0.8, 0.1, 0.7, ...]}. This vector captures subtle meaning dimensions that might include concepts like "transportation," "commercial service," "reservation," and "cancellation policy" without explicitly naming them. Two capabilities with similar embeddings likely have similar semantic meanings, even if they use different terminology.
    
    \item \textbf{Ontological Mapping}: By linking to\\ \texttt{http://schema.org/Flight}, the agent situates itself within a standardized knowledge structure. This enables reasoning engines to understand that this capability is a type of transportation service, which is a type of commercial service, which involves organizations, people, and locations—all without these relationships being explicitly stated in the agent's declaration.
    
    \item \textbf{Relational Potential}: The interface specification defines precisely how other agents can interact with this one. The input/output definitions create a contract: "If you provide me with origin 'SFO', destination 'NRT', departure date '2025-06-15', and 2 passengers in business class, I will return an array of available flight options." This explicit contract forms the basis for reliable multi-agent workflows.
\end{enumerate}

{\mx} addresses embedding standardization through a universal embedding translation mechanism inspired by recent advances in cross-model embedding alignment. Rather than forcing all agents to use identical embedding models, {\mx} leverages the Strong Platonic Representation Hypothesis \cite{jha2025harnessing}, which demonstrates that neural networks trained with similar objectives converge to universal/similar latent spaces that enable translation between different embedding representations without paired data. When agents join the platform, their capabilities are processed through a vec2vec-inspired translation layer \cite{jha2025harnessing} that maps embeddings from different models (e.g., T5-based GTR, BERT-based GTE, or RoBERTa-based Granite) into a shared semantic space, ensuring semantic compatibility while preserving agent autonomy in their choice of embedding models. This prevents compatibility issues where agents using different embedding architectures would otherwise have incompatible semantic representations, enabling easier capability discovery and interaction regardless of each agent's internal embedding choice.

Consider how this differs from traditional API registries that simply list endpoints and parameter types. In a traditional system, a developer must understand what "flightBooking" means through external documentation. In {\mx}, the agent's capability declaration itself provides \textit{both} human-understandable semantics and machine-processable relationships through its ontological commitments.

This theoretical approach aligns with the conceptualization of semantic information as meaningful data that is truthful about its domain \cite{floridi2011philosophy, 2022bayesian}, while providing practical implementation through computational methods.

\subsubsection{Emergent Discovery Through Multi-Modal Matching}

\begin{figure}[t]
\centering
\includegraphics[width=0.35\linewidth,trim=3.5cm 1.5cm 3.5cm 0cm]{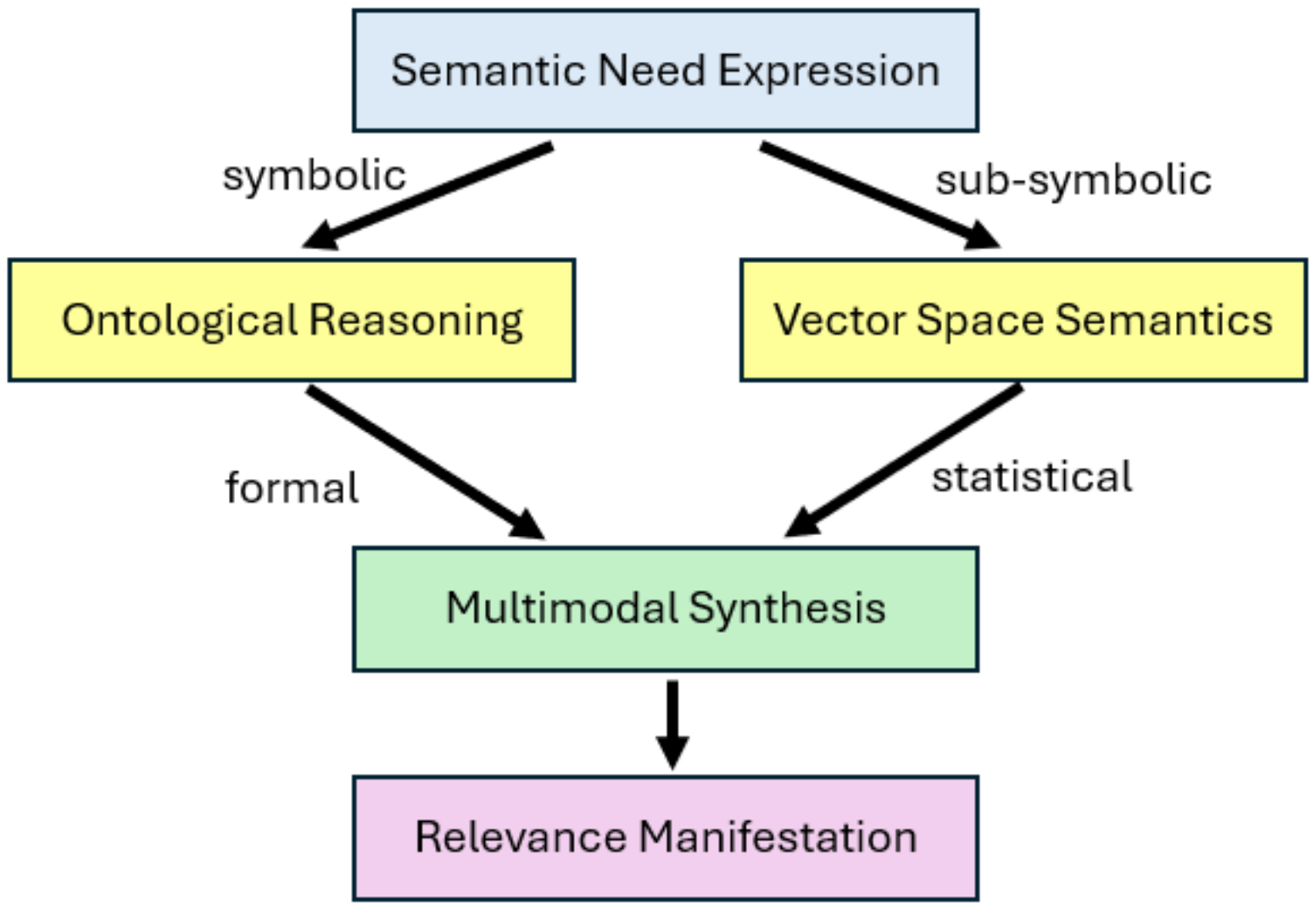}
\caption{The dualism of capability discovery in {\mx}}
\label{fig:discovery-process}
\vspace{-1em}
\end{figure}

The capability discovery process in {\mx} represents a microcosm of how emergent intelligence arises from distributed cognitive systems. As illustrated in Figure~\ref{fig:discovery-process}, the process synthesizes both symbolic and sub-symbolic approaches to knowledge representation.

When an agent articulates a capability need, it transcends mere keyword matching, instead expressing an intentional state about its requirements:

\begin{lstlisting}[caption={Semantic need expression example}, label={lst:discovery}]
{ "required": {
"functionality": "Find and book flights",
"ontology": "travel:Transportation",
"constraints": ["businessClass", "directFlights"] }}
\end{lstlisting}
Let's examine how the discovery process works through a concrete example. Imagine a travel planner agent needs to find a service that can book business-class direct flights. The process unfolds as follows:

\begin{enumerate}
    \item \textbf{Ontological Reasoning (Symbolic Path):} The system examines the ontological relationship between\\ \texttt{travel:Transportation} (requested) and \texttt{schema.org/Flight} (provided by \texttt{flight-agent-001}). Through logical inference, it determines:
    
    \begin{itemize}
        \item \texttt{schema.org/Flight} is a subclass of\\ \texttt{schema.org/TransportService}
        \item \texttt{schema.org/TransportService} is equivalent to\\ \texttt{travel:Transportation}
        \item Therefore, \texttt{flight-agent-001} satisfies the ontological requirement
    \end{itemize}
    
    The system then checks the agent's operations and determines that \texttt{["search", "book"]} operations satisfy the "Find and book flights" functionality requirement.
    
    \item \textbf{Vector Space Semantics (Sub-symbolic Path):} Simultaneously, the system converts "Find and book flights" into a semantic embedding vector—perhaps \texttt{[0.15, 0.79, 0.08, 0.66, ...]}. It then computes the cosine similarity with the flight agent's capability embedding \texttt{[0.2, 0.8, 0.1, 0.7, ...]}, yielding a similarity score of 0.97. This high similarity indicates that the capability semantically aligns with the need, even without understanding the specific words used.
    
    \item \textbf{Constraint Validation:} The system examines the constraints through both symbolic and sub-symbolic methods:
    
    \begin{itemize}
        \item For "businessClass" - the symbolic path easily validates this constraint by finding the explicit "class" parameter in the interface specification.
        
        \item For "directFlights" - though not explicitly present in the interface parameters, the sub-symbolic path handles this constraint. The semantic embedding of the agent's capability contains dimensional components that encode "route directness" features.
        
        \item When the actual agent interaction occurs, this sub-symbolic understanding translates to practical usage in one of two ways:
        
            \begin{itemize}
                \item \textbf{Semantic Translation:} The coordinating agent transforms the abstract "directFlights" constraint into the agent-specific format required by \texttt{flight-agent-001}. For instance, it might add a \texttt{"maxConnections": 0} parameter based on semantic knowledge of how this agent implements the concept of direct flights.
                
                \item \textbf{Post-Processing Filter:} Alternatively, the coordinating agent requests all flights and applies a filter based on connection count in the returned results, utilizing the semantic knowledge that this agent includes connection information in its response format.
            \end{itemize}
            
        \item This practical application of sub-symbolic knowledge demonstrates how semantic understanding extends beyond mere capability discovery to facilitate the actual interaction patterns between agents with different representation systems.
    \end{itemize}
    
    \item \textbf{Multimodal Synthesis:} The system combines evidence from both reasoning pathways: (1) \textbf{Ontological match}: Strong (exact subclass relationship), (2) \textbf{Vector similarity}: Very high (0.97), (3) \textbf{Constraint satisfaction}: Complete (both constraints can be satisfied). The synthesis algorithm weights these factors (perhaps 0.4, 0.4, and 0.2 respectively) to produce a final relevance score of 0.92.
\end{enumerate}

This approach demonstrates how multiple knowledge representation paradigms operate in parallel, creating a more robust discovery mechanism than either could achieve alone. For example, if an agent advertised capabilities using different terminology (e.g., "airTicketing" instead of "flightBooking"), the ontological reasoning might fail to find an exact match, but the vector similarity would still detect the semantic equivalence.

After synthesis, the system manifests discovered capabilities with associated confidence metrics:

\begin{lstlisting}[caption={Discovery manifestation with confidence metrics}, label={lst:results}]
{ "matches": [
{"agentId": "flight-agent-001", "capability": "flightBooking", "score": 0.92},
{"agentId": "travel-agent-005", "capability": "airTicketing", "score": 0.87},
{"agentId": "booking-agent-003", "capability": "reservations", "score": 0.71}
]}
\end{lstlisting}

To further illustrate the robustness of this approach, consider a real-world scenario where ambiguity exists. Imagine a business traveler agent submits a more ambiguous query:

\begin{lstlisting}[caption={Ambiguous capability request}, label={lst:ambiguous}]
{ "required": { "functionality": "arrange executive travel",
"constraints": ["premium", "flexible"] }}
\end{lstlisting}
Without ontological specification, the system must rely heavily on vector similarities and interpretation of constraints.

This demonstrates how the discovery mechanism handles uncertainty—returning a ranked list of possibilities that reflect different interpretations of the ambiguous request. The requesting agent can then select the most appropriate capability based on additional context or user preferences.
The discovery process manifests four (philosophical) principles:
\begin{enumerate}
    \item \textbf{Pluralistic Epistemology}: Knowledge is discovered through multiple complementary pathways. In our example, both the formal ontological reasoning about transportation hierarchies and the statistical similarity of semantic vectors contributed to finding appropriate flight booking services.
    
    \item \textbf{Emergent Relevance}: Meaning emerges from the synthesis of different knowledge representation approaches. The final relevance scores wouldn't be possible from either approach alone—they emerge from the combination of symbolic reasoning and statistical similarity.
    
    \item \textbf{Pragmatic Validation}: The effectiveness of discovered capabilities is determined by their practical utility. While\\ \texttt{flight-agent-001} received the highest score for the specific query, different agents might be more suitable for different contexts. The proof is in the successful execution of the requested task.
    
    \item \textbf{Bounded Rationality}: Agents make satisficing rather than optimal decisions based on incomplete information. The discovery mechanism doesn't guarantee finding the absolute best agent for a task (which would require perfect knowledge of all agents and their capabilities), but rather helps agents find satisfactory solutions with reasonable effort.
\end{enumerate}
This theoretical framework, inspired by cognitive science theories of distributed representation \cite{hinton1986distributed}, ensures that {\mx} can function effectively in open-world settings where complete knowledge is impossible. In this framework, the Universal Message Bus acts not merely as a communication channel but as a manifestation of collective intelligence—a shared cognitive space where agent capabilities can be discovered, evaluated, and combined to solve problems that transcend any individual agent's capabilities. For example, a complete travel itinerary might involve the flight-booking agent, a hotel-reservation agent, a local-transportation agent, and a weather-forecasting agent, all discovered and orchestrated through this capability discovery mechanism. This philosophical perspective aligns with Clark's extended mind thesis \cite{clark1998extended}, viewing the {\mx} ecosystem as an extension of distributed cognitive processes. Just as humans offload cognitive tasks to tools and other people, agents in {\mx} offload capability discovery to the collective infrastructure, extending their effective capabilities beyond their individual programming.

\subsection{Reconciling Autonomy with Context: Contextual State Sharing}

A critical tension exists between A2A's commitment to statelessness and MOD-X's distributed state approach. A2A embraces statelessness as a fundamental design principle, ensuring agent independence, simplified reasoning, and clear security boundaries. However, effective communication often depends on shared context that transcends what can be practically included in each message, particularly for complex interactions. MOD-X resolves this tension through "contextual state sharing"—a middle ground that preserves agent autonomy while enabling rich collaborative interactions:

\begin{lstlisting}[caption={Contextual state sharing}, label={lst:contextual-state}]
// Agent declares its state autonomy model
{ "agentId": "flight-agent-001",
"statePolicy": {
"defaultMode": "stateless",  // Presumes autonomy
"contextualSharing": {
"enabled": true,
"contexts": ["travelPlanning", "emergencyRerouting"],
"shareableStateTypes": ["flightAvailability", "pricingData"],
"stateLifespan": "task-bounded",
"revocable": true }}}
\end{lstlisting}
This approach allows agents to form shared commitments and intentions while maintaining their individual agency. Agents in MOD-X maintain full autonomy by default but can explicitly opt into shared contexts when collaboration requires it—similar to how teams form temporarily for specific projects while members maintain their independence. 

The contextual sharing model implements this approach through several practical mechanisms: The contextual sharing model implements this approach through several practical mechanisms: \textbf{Explicit Consent}—agents must actively opt into shared contexts rather than having state imposed upon them; \textbf{Bounded Scope}—shared state is limited to specific contextual domains relevant to collaboration; \textbf{Temporal Boundaries}—context sharing has explicit lifespans, typically bounded by task completion; and \textbf{Revocation Rights}—\\agents maintain the right to withdraw from shared contexts.

This middle-ground approach enables MOD-X to support sophisticated collaborative behaviors while respecting the commitment to agent autonomy. When applied to our travel planning scenario, this means the flight booking agent can temporarily share its knowledge of available flights and pricing within the specific context of planning a Tokyo business trip, while maintaining its autonomy and not exposing unrelated state or capabilities. Once the trip is booked, this contextual sharing expires, returning all agents to their default stateless interaction mode.

\subsection{Security and Trust}

{\mx}'s blockchain-based security layer provides stronger guarantees for agent identity and interaction verification than the OAuth/JWT approaches used by A2A and MCP. By creating immutable records of agent interactions and implementing reputation systems, {\mx} addresses the trust challenges inherent in decentralized agent ecosystems.

\section{Worked Example: Travel Planning System}

To illustrate {\mx}'s practical application, we present a concrete example involving a travel planning system with multiple specialized agents.
A user needs to plan a business trip to Tokyo, requiring flight bookings, accommodations, local transportation, and calendar coordination, while staying within a \$3,000 budget. The scenario involves several specialized agents: The scenario involves several specialized agents: \textbf{FlightBookingAgent}—searches and books flights; \textbf{AccommodationAgent}—recommends hotels; \textbf{LocalTransportAgent}—provides transportation; \textbf{CalendarAgent}—manages user's schedule; \textbf{BudgetAgent}—enforces budget constraints; and \textbf{CoordinatorAgent}—orchestrates workflow.

\subsubsection*{Capability Discovery}

When a request is submitted, the CoordinatorAgent queries for required capabilities:

\begin{small}
\begin{verbatim}
// CoordinatorAgent Query
{ "messageType": "CapabilityQuery",
"capabilities": ["flightBooking", "accommodation",
"localTransport", "calendar",
"budgetManagement"],
"requestId": "trip-planning-12345" }
\end{verbatim}
\end{small}

\subsubsection*{Message Exchange Pattern}

\begin{figure}[t]
\centering
\includegraphics[width=0.35\linewidth, trim=2.2cm 2.7cm 2.3cm 2cm]{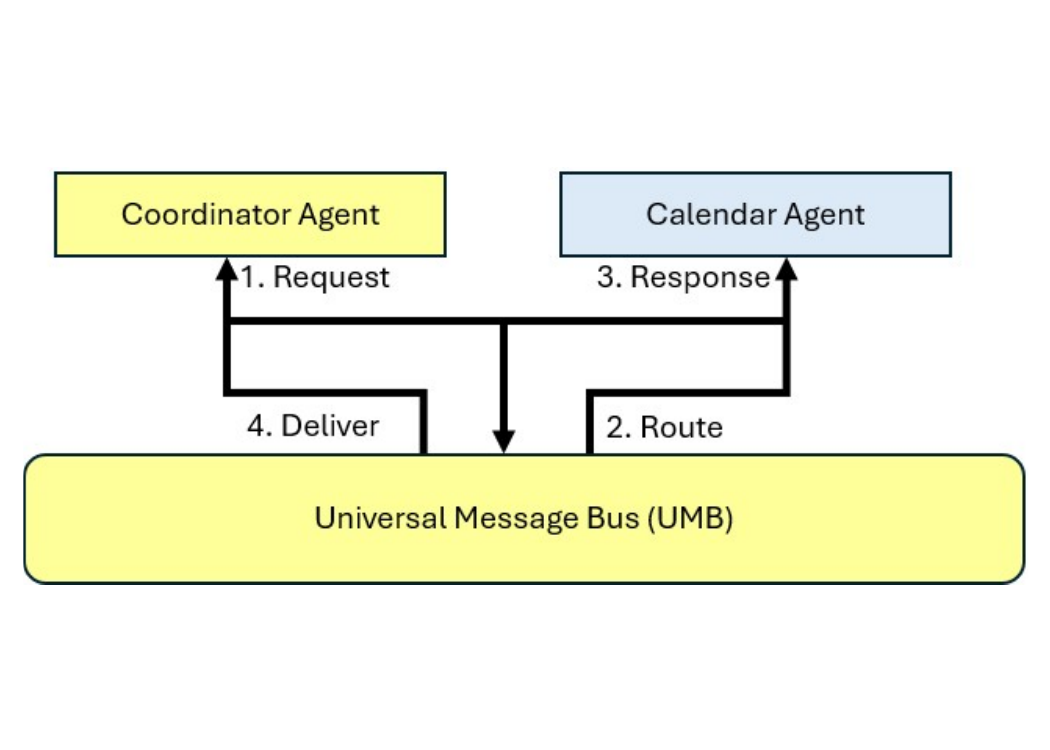}
\caption{Agent message exchange through the Universal Message Bus}
\label{fig:message-exchange}
\end{figure}

The Universal Message Bus (UMB) facilitates communication between agents using a standardized message routing pattern. As shown in Figure \ref{fig:message-exchange}, messages flow through the UMB rather than directly between agents, enabling decentralized many-to-many communications: (1) request from Coordinator, (2) routing to Calendar Agent, (3) response from Calendar Agent, and (4) delivery back to Coordinator. This approach differs from the client-server patterns used in A2A and MCP protocols, providing greater flexibility and fault tolerance.

\subsubsection*{Data Transformation and Semantic Matching}

\begin{figure}[t]
\centering
\begin{tikzpicture}[
    scale=0.8, transform shape,
    box/.style={draw, text width=3.5cm, align=left, minimum height=4.5cm, font=\tiny},
    arrow/.style={->, thick},
    trans/.style={draw, align=center, text width=2.5cm, font=\scriptsize}
]
    
    \node[box] (original) at (-1,0) {
        \textbf{Flight Agent (Airline Ontology):}
        \vspace{0.1cm}
        
        \texttt{\{
          "flightOptions": [{\\
            "carrier": "ANA",\\
            "flightNo": "NH007",\\
            "departure": \{\\
              "airport": "SFO",\\
              "time": "2025-06-10T10:30Z"\\
            \},\\
            "arrival": \{\\
              "airport": "NRT",\\
              "time": "2025-06-11T14:25Z"\\
            \},\\
            "price": 1650,\\
            "class": "business"\\
          }]
        \}}
    };

    \node[box] (standard) at (6,0) {
        \textbf{Budget Agent (Travel Ontology):}
        \vspace{0.1cm}
        
        \texttt{\{
          "travelSegments": [{\\
            "type": "flight",\\
            "provider": "ANA",\\
            "identifier": "NH007",\\
            "origin": \{\\
              "location": "San Francisco",\\
              "departure": "2025-06-10T10:30Z"\\
            \},\\
            "destination": \{\\
              "location": "Tokyo",\\
              "arrival": "2025-06-11T14:25Z"\\
            \},\\
            "cost": 1650,\\
            "category": "premium"\\
          }]
        \}}
    };

    \node[trans] (translation) at (2.5,0) {Translation Layer\\
    \scriptsize{
    1. Format conversion\\
    2. Semantic mapping\\
    3. Ontology alignment
    }};

    \draw[arrow] (original) -- (translation);
    \draw[arrow] (translation) -- (standard);
\end{tikzpicture}
\caption{Data transformation and semantic matching by the Translation Layer}
\label{fig:translation}
\end{figure}
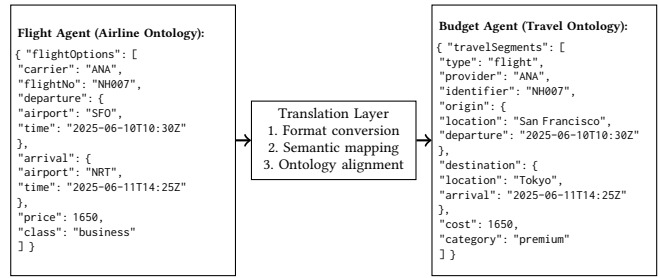

A key innovation in {\mx} is the Translation Layer, which enables interoperability between heterogeneous agents with different message formats and semantic concepts. Figure \ref{fig:translation} demonstrates this dual capability.

\noindent The translation process performs several key functions: The translation process performs several key functions: \textbf{Format Conversion}—transforming the structure from flight-specific to generic travel format; \textbf{Semantic Mapping}—converting between conceptual equivalents (e.g., "class": "business" → "category": "premium"); \textbf{Ontology Alignment}—resolving differences in domain conceptualizations (e.g., airport code "SFO" to city name "San Francisco"); and \textbf{Contextual Enhancement}—adding implied information not explicitly stated in the original message.

This semantic interoperability enables agents built on entirely different knowledge representations to communicate effectively. For example, the Flight Agent uses an airline industry ontology with airport codes, while the Budget Agent uses a travel planning ontology with city names. The {\mx} Translation Layer maintains knowledge of concepts that enables these different domain models to be automatically aligned and transformed at runtime.

\subsubsection*{Security Verification}

Following {\mx}'s hybrid security model, this high-value flight booking transaction uses blockchain verification to create an immutable record:

\begin{small}
\begin{verbatim}
// Transaction Verification Record (High-Value Operation)
{ "transactionType": "flightBooking",
"agentId": "flight-agent-001",
"requestorId": "coordinator-agent-main",
"timestamp": "2025-05-17T09:42:17Z",
"actionParameters": { "flight": "NH007",
"passenger": "user-12345",
"cost": 1650 },
"verificationStatus": "approved",
"securityToken": "eyJhbGciOiJFZER..." }
\end{verbatim}
\end{small}

In contrast, routine operations (e.g. capability queries, status updates) should use lightweight cryptographic signatures for performance without blockchain overhead.

\section{Limitations}
While {\mx} represents a significant advancement over existing protocols, key challenges remain: developing formal specifications for widespread adoption, evaluating performance at scale, and establishing decentralized governance mechanisms.

\section{Conclusion}
This paper introduces {\mx} (Modular Open Decentralized eXchange), an architectural framework designed to enable integration of heterogeneous specialist AI agents for better Information Retrieval. {\mx} addresses key limitations of existing protocols through its layered architecture, which includes a Universal Message Bus, Translation Layer, State Management, and Security mechanisms. {\mx}'s distinguishing features include a truly decentralized publish-subscribe communication model that enables many-to-many agent interactions without central coordination, semantic capability discovery that allows agents to find appropriate collaborators based on required capabilities, elaborate state management that allows maintaining consistent context across distributed agent systems, blockchain-based security that provides strong guarantees for agent identity and interaction verification, and a dynamic workflow engine that orchestrates complex multi-agent collaborations.

\newpage
\bibliographystyle{ACM-Reference-Format}
\bibliography{sigconf}

\end{document}